\documentclass[11pt,a4paper]{article}
\usepackage[hyperref]{acl2020}
\usepackage{times}
\usepackage{latexsym}
\usepackage{tabularx}
\usepackage{graphicx}
\usepackage{url}
\usepackage{amsmath}
\usepackage{mathtools}
\usepackage{amsfonts}
\usepackage{bm}
\usepackage{bbold}
\usepackage{booktabs}  
\usepackage{comment}
\usepackage{algorithm}
\usepackage{algorithmic}
\usepackage{pifont}

\newcommand{\xmark}{\ding{55}}
\newcommand{\quotes}[1]{``#1''}

\usepackage{xcolor}

\aclfinalcopy 

\title{Towards Faithful Neural Table-to-Text Generation with \\ Content-Matching Constraints}

\author{
  Zhenyi Wang$^{\dagger\mathsection}$\thanks{Zhenyi Wang was a research intern student at Tencent AI Lab in Bellevue, WA when doing this work.}, Xiaoyang Wang$^{\mathsection}$, Bang An$^{\dagger}$,  Dong Yu$^{\mathsection}$, Changyou Chen$^{\dagger}$\\
  $^{\dagger}$State University of New York at Buffalo, $^{\mathsection}$Tencent AI Lab, Bellevue, WA\\
  \texttt{\{zhenyiwa, anbang, changyou\}@buffalo.edu}\\
  \texttt{\{shawnxywang, dyu\}@tencent.com}
}

\date{}

\begin{document}
\maketitle
\begin{abstract}
Text generation from a knowledge base aims to translate knowledge triples to natural-language descriptions. Most existing methods ignore the faithfulness between a generated text description and the original table, leading to generated information that goes beyond the content of the table. In this paper, for the first time, we propose a novel Transformer-based generation framework to achieve the goal. The core techniques in our method to enforce faithfulness include a new table-text optimal-transport matching loss and a table-text embedding similarity loss based on the Transformer model. Furthermore, to evaluate faithfulness, we propose a new automatic metric specialized to the table-to-text generation problem. We also provide detailed analysis on each component of our model in our experiments. Automatic and human evaluations show that our framework can significantly outperform state-of-the-art by a large margin.

\end{abstract}

\section{Introduction}
\begin{figure}[t]
    \centering
    \includegraphics[width=0.95\linewidth]{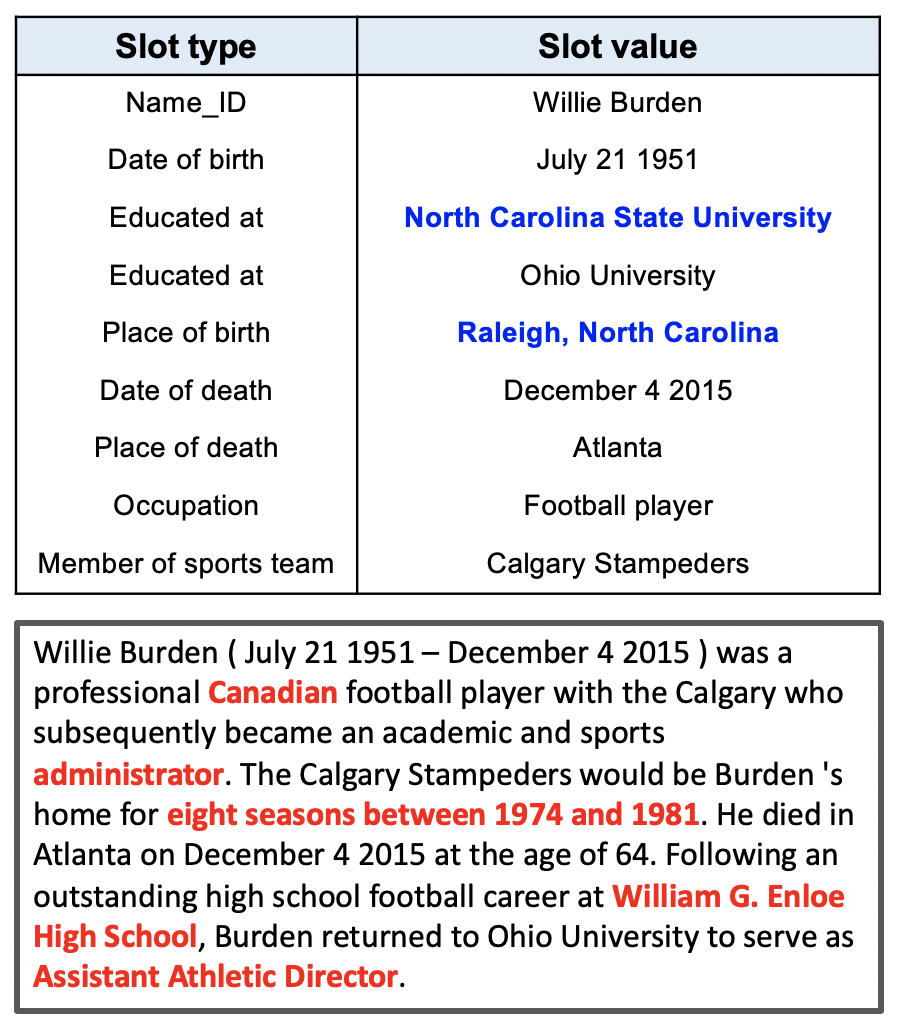}
    \caption{An example of table-to-text generation. This generation is unfaithful because there exists information in table not covered by generated text (marked in blue); At the same time, hallucinated information in text does not appear in table (marked in red).}
    \label{fig:example}
    \vspace{-0.5cm}
\end{figure}
Understanding structured knowledge, \textit{e.g.}, information encoded in tables, and automatically generating natural-language descriptions is an important task in the area of Natural Language Generation. 
Table-to-text generation helps making knowledge elements and their connections in tables easier to comprehend by human. 
There have been a number of practical application scenarios in this field, for example, weather report generation, NBA news generation, biography generation and medical-record description generation \cite{weather, NBA, bio, medical}.

Most existing methods for table-to-text generation are based on an encoder-decoder
framework \cite{Sutskever2014,Bahdanau2015}, most of which are RNN-based Sequence-to-Sequence (Seq2Seq) models \cite{Wikiemnlp2016,structure2018,Wiseman2018,Ma2019,Wang2018,tianyu2019}. 
Though significant progress has been achieved, we advocate two key problems in existing methods. 
Firstly, because of the intrinsic shortage of RNN, RNN-based models are not able to capture long-term dependencies, 
which would lose important information reflected in a table. 
This drawback prevents them from being applied to larger tables, for example, a table describing a large Knowledge Base \cite{Wang2018}. Secondly, little work has focused on generating faithful text descriptions, which is defined, in this paper, as the level of matching between \textit{a generated text sequence and the corresponding table content}. An unfaithful generation example is illustrated in Figure \ref{fig:example}. The training objectives and evaluation metrics of existing methods encourage generating texts to be as similar as possible to reference texts. One problem with this is that the reference text often contains extra information that is not presented in the table because human beings have external knowledge beyond the input table when writing the text, or it even misses some important information in the table \cite{Dhingra2019} due to the noise from the dataset collection process. As a result, unconstrained training with such mis-matching information usually leads to hallucinated words or phrases in generated texts, making them unfaithful to the table and thus harmful in practical uses.

In this paper, we aim to overcome the above problems to automatically generate faithful texts from tables. In other words, we aim to produce the writing that a human without any external knowledge would do given the same table data as input.
In contrast to existing RNN-based models, we leverage the powerful attention-based Transformer model to capture long-term dependencies and generate more informative paragraph-level texts.
To generate descriptions faithful to tables, two content-matching constraints are proposed. 
The first one is a latent-representation-level matching constraint encouraging the latent semantics of the whole text to be consistent with that of the whole table. 
The second one is an explicit entity-level matching scheme, which utilizes Optimal-Transport (OT) techniques to constrain key words of a table and the corresponding text to be as identical as possible. 
To evaluate the faithfulness, we also propose a new PARENT-T metric evaluating the content matching between texts and tables, based on the recently proposed PARENT \cite{Dhingra2019} metric. 
We train and evaluate our model on a large-scale knowledge base dataset \cite{Wang2018}. 
Automatic and human evaluations both show that our method achieve the state-of-the-art performance, and can generates paragraph-level descriptions much more informative and faithful to input tables.

\begin{figure}[!t]
    \centering
    \includegraphics[width=\linewidth]{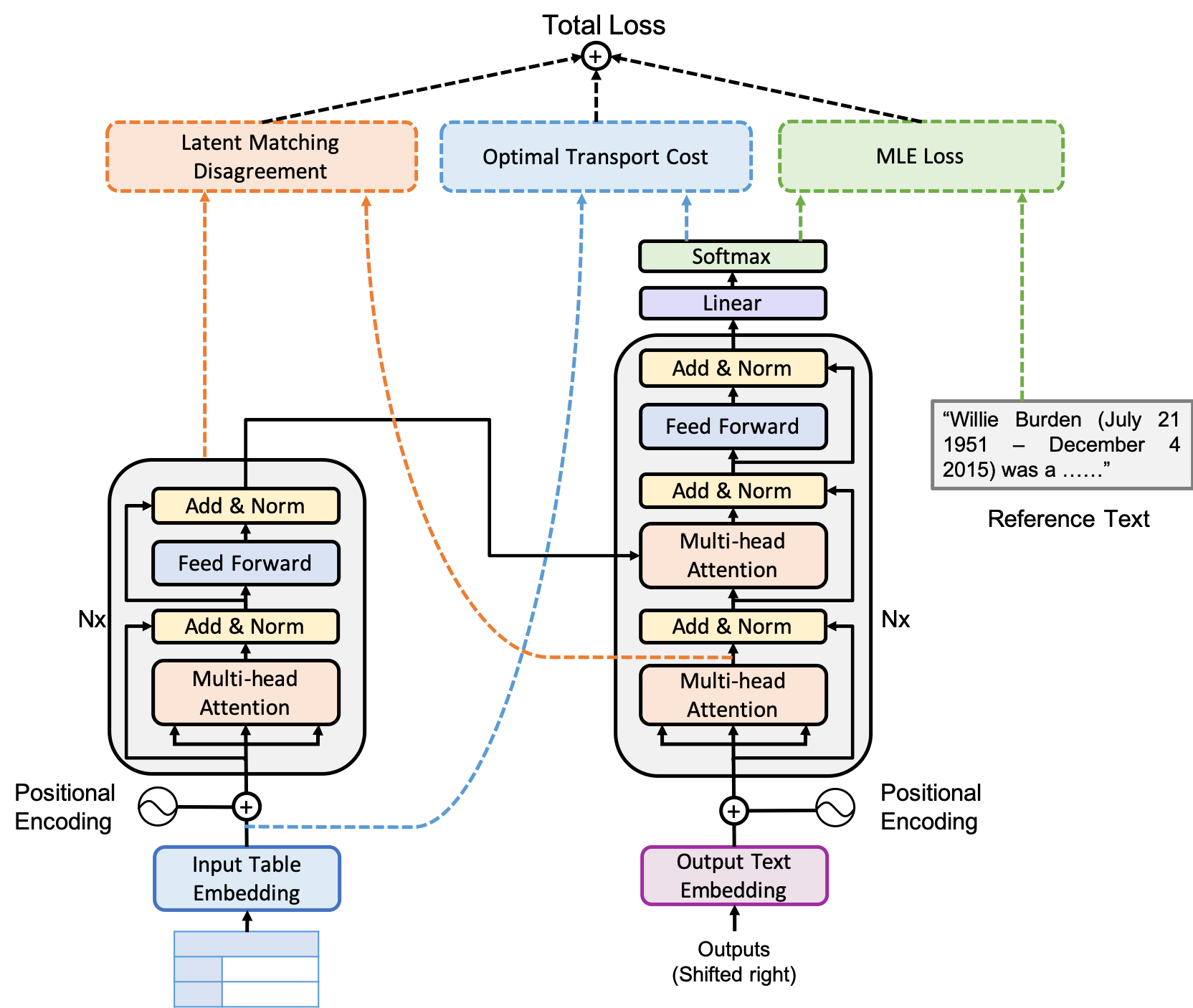}
    \vspace{-0.6cm}
    \caption{The architecture of our proposed model for table-to-text generation. To enhance the ability of generating multi-sentence faithful texts, our loss consists of three parts, including a maximum-likelihood loss (green), a latent matching disagreement loss (orange), and an optimal-transport loss (blue).}
    \label{fig:architecture}
    \vspace{-0.5cm}
\end{figure}

\section{The Proposed Method}

The task of text generation for a knowledge base is to take the structured table, $\bm{T} = \{(t_1, v_1),(t_2, v_2), · · · ,(t_m, v_m) \}$, as
input, and outputs a natural-language description consisting of a sequence of words $\bm{y} =
\{y_1, y_2, · · · , y_n\}$ that is faithful to the input table. Here, $t_i$ denotes the slot type for the $i^{th}$ row, and $v_i$ denotes the slot value for the $i^{th}$ row in a table.

Our model adopts the powerful Transformer model \cite{Vaswani2017} to translate a table to a text sequence. Specifically, the Transformer is a Seq2Seq model, consisting of an encoder and a decoder. Our proposed encoder-to-decoder Transformer model learns to estimate the conditional probability of a text sequence from a source table input in an autoregressive way:
\vspace{-0.3cm}
\begin{equation} \label{eq:mle}
    P(\bm{y}|\bm{T}; \bm{\theta}) = \prod_{i=1}^n P(y_i|\bm{y}_{< i}, \bm{T}; \bm{\theta})~,
\end{equation}\par\vspace{-0.2cm}
\noindent where $\bm{\theta}$ is the Transformer parameters and $\bm{y}_{< i}$ denotes the decoded words from previous steps. 

Existing models for table-to-text generation either only focus on generating text to match the reference text \cite{structure2018, Ma2019}, or only require a generated text sequence to be able to cover the input table \cite{Wang2018}. However, as the only input information is the table, the generated text should be faithful to the input table as much as possible. Therefore, we propose two constraint losses, including a table-text disagreement constraint loss and a constrained content matching loss with optimal transport, to encourage the model to learn to match between the generated text and the input table faithfully. Figure~\ref{fig:architecture} illustrates the overall architecture of our model. In summary, our model loss contains three parts: 1) a maximum likelihood loss (green) that measures the matching between a model prediction and the reference text sequence; 2) a latent feature matching disagreement loss (orange) that measures the disagreement between a table encoding and the corresponding reference-text encoding; and 3) an optimal-transport loss (blue) matching the key words of an input table and the corresponding generated text.

\subsection{Table Representation}
The entities of a table simply consists of Slot Type and Slot Value pairs. 
To apply the Transformer model, we first linearize input tables into sequences. Slot types and slot values are separated by special tokens  ``$<$'' and ``$>$''. As an example, the table in Figure \ref{fig:example} is converted into a sequence: $\{<\text{Name\_ID}>, \text{Willie Burden}, <\text{date of birth}>, \text{July 21 1951}, \cdots \}$. 
We note that encoding a table in this way might lose some high-order structure information presented in the original knowledge graph. However, our knowledge graph is relatively simple. According to our preliminary studies, a naive combination of feature extracted with graph neural networks \cite{beck2018} does not seem helpful. 
As a result, we only rely on the sequence representation in this paper.

\subsection{The Base Objective}
Our base objective comes from the standard Transformer model, which is defined as the negative log-likelihood loss $\mathcal{L}_{mle}$ of a target sentence $\bm{y}$ given its input $\bm{T}$, {\it i.e.},
\begin{equation}
    \mathcal{L}_{mle} =  -\log  P(\bm{y}|\bm{T}; \bm{\theta}) 
\end{equation}
with $P(\bm{y}|\bm{T}; \bm{\theta})$ defined in \eqref{eq:mle}.

\subsection {Faithfulness Modeling with a Table-Text Disagreement Constraint Loss}

One key element of our model is to enforce a generated text sequence to be consistent with (or faithful to) the table input. To achieve this, we propose to add some constraints so that a generated text sequence only contains information from the table. Our first idea is inspired by related work in machine translation \cite{sentagree2019}. Specifically, we propose to constrain a table embedding to be close to the corresponding target sentence embedding. Since the embedding of a text sequence (or the table) in our model is also represented as a sequence, we propose to match the mean embeddings of both sequences. In fact, the mean embedding has been proved to be an effective representation for the whole sequence in machine translation \cite{sentagree2019, wang2017}. Let $\hat{\bm{V}}_{\text{table}}$ and $\hat{\bm{V}}_{\text{text}}$ be the mean embeddings of a table and the target text embeddings in our Transformer-based model, respectively. A table-target sentence
disagreement loss $\mathcal{L}_{\text{disagree}}$ is then defined as
\begin{equation}
    \mathcal{L}_{\text{disagree}} =  \lVert \hat{\bm{V}}_{\text{table}} - \hat{\bm{V}}_{\text{text}} \rVert ^2
\end{equation}

\begin{figure*}
    \centering
    
    \begin{minipage}{0.49\linewidth}
        \includegraphics[width=\linewidth]{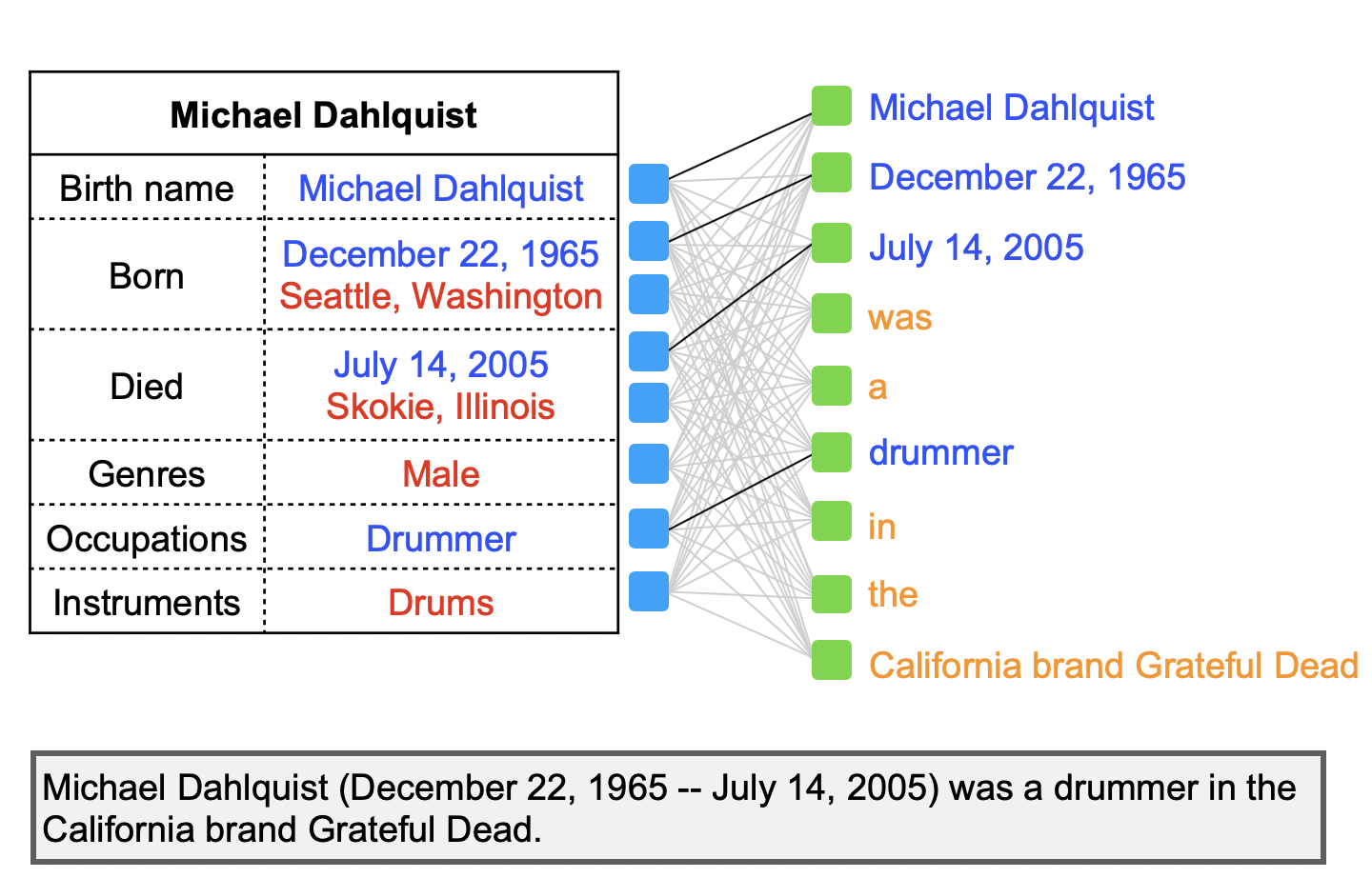}
    \end{minipage}
    \begin{minipage}{0.49\linewidth}
        \includegraphics[width=\linewidth]{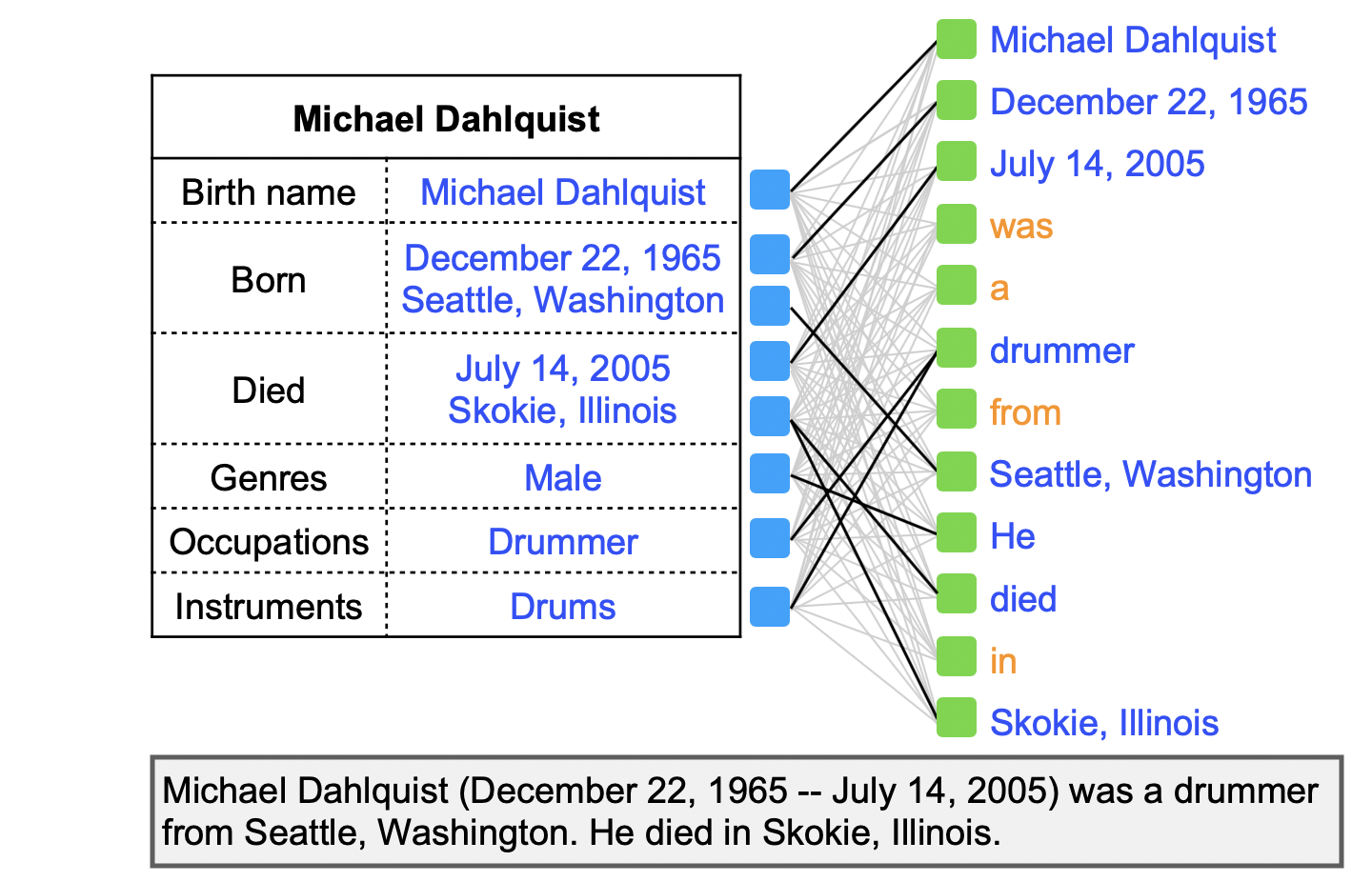}
    \end{minipage}
    
    \caption{Illustration of the OT loss, which is defined with OT distance to only match key words in both the table and the generated sentence. Left: the generated sentence not only contains extra information not presented in the table (shown as orange), but also lacks some information presented in the table (shown as red).  This is unfaithful generation. The OT lost is thus high. Right: all information in the table is covered in the generated sentence, and the generated sentence does not contain extra information not presented in the table. This is faithful generation. The OT cost is thus low. This example is borrowed and modified from \cite{Dhingra2019}.}
    \label{fig:OT_loss}
    \vspace{-0.4cm}
\end{figure*}

\subsection {Faithfulness Modeling with Constrained Content Matching via Optimal Transport} \label{sec:OT}
Our second strategy is to explicitly match the key words in a table and the corresponding generated text. In our case, key words are defined as nouns, which can be easily extracted with existing tools such as NLTK \cite{loper2002}. To match key words, a mis-matching loss should be defined. Such a mis-matching loss could be non-differentiable, {\it e.g.}, when the loss is defined as the number of matched entities. In order to still be able to learn by gradient descent, one can adopt the policy gradient algorithm to deal with the non-differentiability. However, policy gradient is known to exhibit high variance. To overcome this issue, we instead propose to perform optimization via optimal transport (OT), inspired by the recent techniques in \cite{liqun2019}.

\paragraph{Optimal-Transport Distance}
In the context of text generation, a generated text sequence, $\bm{y} = (y_1, \cdots, y_n)$, can be represented as a discrete distribution $\bm{\mu} = \sum_{i=1}^n u_i\delta_{y_i}(\cdot)$, where $u_i \geq 0$ and $\sum_iu_i = 1$, $\delta_x(\cdot)$ denotes a spike distribution located at $x$. Given two discrete distributions $\bm{\mu}$ and $\bm{\nu}$, written as $\bm{\mu} = \sum_{i=1} ^ {n} \bm{u}_{i}  \delta_{\bm{x}_i}$ and $\bm{\nu} = \sum_{j=1} ^ {m} \bm{v}_{j}  \delta_{\bm{y}_j}$, respectively, the OT distance between $\bm{\mu}$ and $\bm{\nu}$ is defined as the solution of the following maximum network-flow problem: 
\vspace{-0.3cm}
\begin{equation}\label{eq:OTloss}
 \mathcal{L}_{\text{OT}} =  \underset {\bm{U}  \in \Pi(\bm{\mu}, \bm{\nu})} {\min}  \sum_{i=1} ^{n} \sum_{j=1} ^{m} \bm{U}_{ij} \cdot d(\bm{x}_i,\bm{y}_j)~,
\end{equation}\par\vspace{-0.3cm}
\noindent where $d(\mathbf{x},\mathbf{y})$ is the cost of moving $\mathbf{x}$ to $\mathbf{y}$ (matching $\mathbf{x}$ and $\mathbf{y}$). In this paper, we use the cosine distance between the two word-embedding vectors of $\mathbf{x}$ and $\mathbf{y}$, defined as $d(\mathbf{x},\mathbf{y})= 1- \frac{\mathbf{x} \mathbf{y}}{{\rVert \mathbf{x} \rVert}_2  {\rVert \mathbf{y} \rVert}_2}$. $\Pi(\bm{\mu}, \bm{\nu})$ is the set of joint distributions such that the two marginal distributions equal to $\bm{\mu}$ and $\bm{\nu}$, respectively.

Exact minimization over $\bm{U}$ in the above problem is in general computational
intractable \cite{Genevay2018}. Therefore, we adopt the recently proposed Inexact Proximal point method for Optimal Transport (IPOT) \cite{yujia2018} as an approximation. The details of the IPOT algorithm are shown in Appendix \ref{app:IPOT}.

\paragraph{Constrained Content Matching via OT}
To apply the OT distance to our setting, we need to first specify the atoms in the discrete distributions. Since nouns typically are more informative, we propose to match the nouns in both an input table and the decoded target sequence. We use NLTK~\cite{loper2002} to extract the nouns that are then used for computing the OT loss. In this way, the computational cost can also be significantly reduced comparing to matching all words. 

The OT loss can be used as a metric to measure the goodness of the match between two sequences. To illustrate the motivation of applying the OT loss to our setting, we provide an example illustrated in Figure \ref{fig:OT_loss}, where we try to match the table with the two generated text sequences. On the left plot, the generated text sequence contains ``California brand Grateful Dead'', which is not presented in the input table. Similarly, and the phrases "Seattle, Washington" and ``Skokie Illinois'' in the table are not covered by the generated text. Consequently, the resulting OT loss will be high. By contrast, on the right plot, the table contains all information in the text, and all the phrases in the table are also covered well by the generated text, leading to a low OT loss. As a result, optimizing over the OT loss in \eqref{eq:OTloss} would enforce faithful matching between a table and its generated text.

\paragraph{Optimization via OT}
When optimizing the OT loss with the IPOT algorithm, the gradients of the OT loss is required to be able to propagate back to the Transformer component. In other words, this requires gradients to flow back from a generated sentence. Note that a sentence is generated by sampling from a multinomial distribution, whose parameter is the Transformer decoder output represented as a logit vector $\bm{S}_t$ for each word in the vocabulary. This sampling process is unfortunately non-differentiable. To enable back-propagation, we follow \citet{liqun2019} and use the Soft-argmax trick to approximate each word with the corresponding soft-max output.

To further reduce the number of parameters and improve the computational efficiency, we adopt the {\em factorized embedding parameterization} proposed recently \cite{ALBERT2019}. Specifically, we decompose a word embedding matrix of size $V\times D$ into the product of two matrices of sizes $V\times H$ and $H\times D$, respectively. In this way, the parameter number of the embedding matrices could be significantly reduced as long as $H$ is to be much smaller than $D$.

\subsection{The Final Objective}

Combing all the above components, the final training loss of our model is defined as:
\vspace{-0.2cm}
\begin{equation}
\mathcal{L} = \mathcal{L}_{mle} + \lambda \mathcal{L}_{disagree} +  \gamma \mathcal{L}_{OT}~,
\end{equation}\par\vspace{-0.2cm}
\noindent where $\lambda$ and $\gamma$ controls the relative importance of each component of the loss function. 

\subsection {Decoder with a Copy Mechanism}
To enforce a generated sentence to stick to the words presented in the table as much as possible, we follow
\cite{see2017} to employ a copy mechanism when generating an output sequence. Specifically, let $P_{\text{vocab}}$ be the output of the Transformer decoder. $P_{\text{vocab}}$ is a discrete distribution over the vocabulary words and denotes the probabilities of generating the next word. The standard methods typically generate the next word by directly sampling from $P_{\text{vocab}}$. In the copy mechanism, we instead generate the next word $y_i$ with the following discrete distribution:
\begin{align*}
    P(y_i) = p_gP_{\text{vocab}}(y_i) + (1 - p_g)P_{\text{att}}(y_i)~,
\end{align*}
where $p_{g} = \sigma (\bm{W}_1 \bm{h}_i +\bm{b}_1)$ is the probability of switching sampling between $P_{\text{vocab}}$ and $P_{\text{att}}$, with learnable parameters $(\bm{W}_1, \bm{b}_1)$ and $\bm{h}_i$ as the hidden state from the Transformer decoder for the $i$-th word. $P_{\text{att}}$ is the attention weights (probability) returned from the encoder-decoder attention module in the Transformer. Specifically, when generating the current word $y_i$, the encoder-decoder attention module calculates the probability vector $P_{\text{att}}$ denoting the probabilities of attending to each word in the input table. Note that the probabilities of the words not presented in the table are set to zero.

\section{Experiments}
We conduct experiments to verify the effectiveness and superiority of our proposed approach against related methods.

\subsection{Dataset}

Our model is evaluated on the large-scale knowledge-base Wikiperson dataset released by \citet{Wang2018}. It contains 250,186, 30,487, and 29,982 table-text pairs for training, validation, and testing, respectively. 
Compared to the WikiBio dataset used in previous studies \cite{Wikiemnlp2016,structure2018,Wiseman2018,Ma2019} whose reference text only contains one-sentence descriptions, this
dataset contains multiple sentences for each table to cover as many facts encoded in the input structured knowledge base as possible.

\subsection{Evaluation Metrics}

For automatic evaluation, we apply the widely used evaluation metrics including the standard BLEU-4 \cite{Papineni2002}, METEOR \cite{Denkowski2014} and ROUGE \cite{Lin2004} scores to evaluate the generation quality. 
Since these metrics rely solely on the reference texts, they usually show poor correlations with human judgments when the references deviate too much from the table. 
To this end, we also apply the PARENT \cite{Dhingra2019} metric that considers both the reference texts and table content in evaluations. 
To evaluate the faithfulness of the generated texts, we further modify the PARENT metric to measure the level of matching between generated texts and the corresponding tables. We denote this new metric as PARENT-T. Please see Appendix \ref{app:parent-t} for details. Note that the precision in PARENT-T corresponds to the percentage of words in a text sequence that co-occur in the table; and the recall corresponds to the percentage of words in a table that co-occur in the text.

\begin{table*}[t!]
\centering
\scalebox{0.8}{
  \begin{tabular}{ c c c c c c } \toprule
     \centering
    &BLEU &METEOR &ROUGE &PARENT &PARENT-T  \\  \hline
    \cite{Wang2018} &16.20 & 19.01 & 40.10& 51.03 &54.22 \\
    Seq2Seq \cite{Bahdanau2015} &22.24 & 19.50 & 39.49 & 43.41& 44.55 \\ 
    Pointer-Generator \cite{see2017} &19.32&19.88&40.68&49.52 &52.62 \\
   Structure-Aware Seq2Seq \cite{structure2018}&22.76&20.27&39.32& 46.47 & 48.47\\ \hline
   Ours &\textbf{24.56}&\textbf{22.37}&\textbf{42.40}&\textbf{53.06}&\textbf{56.10}\\
    \hline
  \end{tabular}
  }
  \caption{Comparison of our model and baseline. PARENT and PARENT-T are the \textbf{average} of PARENT and PARENT-T scores of all table-text pairs.}
  \label{tab:baseline}
\end{table*}

\begin{table*}[t!]
\centering
\scalebox{0.8}{
  \begin{tabular}{cccccccc} \toprule
     \centering
    & P-recall & P-precision & PT-recall &  PT-precision\\ \midrule
    \cite{Wang2018} &44.83 &\textbf{63.92} &84.34 &41.10\\
    Seq2Seq \cite{Bahdanau2015} &41.80 &49.09 &76.07 & 33.13\\ 
    Pointer-Generator \cite{see2017} &44.09 &61.73&81.65&42.03\\ 
   Structure-Aware Seq2Seq \cite{structure2018}& 46.34&51.18 &83.84 &35.99\\ \hline
 
    Ours &\textbf{48.83} &62.86&\textbf{85.21}&\textbf{43.52}\\
    \hline
  \end{tabular}
  }
  \caption{Comparison of our model and baseline. P-recall and P-precision refer to the \textbf{average} of PARENT precisions and recalls of all table-text pairs. Similarly, PT-recall and PT-precision are the \textbf{average} of PARENT-T precisions and recalls of all table-text pairs.}
  \label{tab:PR}
\end{table*}

\begin{table*}[t!]
\centering
\scalebox{0.8}{
  \begin{tabular}{cccccccccccc}
  \toprule  
     \centering
 Copy &EF & OT (N/W) & latent &BLEU &METEOR &ROUGE &PARENT &PARENT-T & params\\ \midrule

 \xmark &\xmark & \xmark & \xmark&24.49 &22.01 &40.98 &48.31 &49.89 &98.92M\\ 
 \checkmark  &\xmark &\xmark & \xmark &24.57 &22.43 &42.26 &51.87 &54.29 &98.92M\\ 
    \hline
\checkmark &\checkmark&\xmark &\xmark &25.07&22.38&42.37&51.76&54.36&45.94M\\
\checkmark &\checkmark &\xmark  &\checkmark &23.86&22.08&42.65&52.72&55.30&45.94M\\
 \checkmark &\checkmark & W& \xmark  &24.64&22.39 &42.52&52.77&55.46&45.94M\\
\checkmark &\checkmark& N& \xmark   &25.29&22.60&42.25&52.74&55.80& 45.94M\\
\checkmark  &\checkmark  & N& \checkmark  &24.56&22.37&42.40&53.06&56.10& 45.94M\\
    
    \bottomrule
  \end{tabular}
  }
  \caption{Ablation study of our model components. \checkmark means the corresponding column component is used. \xmark means do not use the corresponding column component. Specifically, \quotes{Copy} means using copy mechanism, \quotes{EF} means using embedding factorization, \quotes{OT} means using optimal transport constraint loss, \quotes{N} means extracting nouns from both the table and text, and \quotes{W} means using the whole table and text to compute OT. Lastly, \quotes{latent} means using latent similarity loss. }
  \label{tab:ablation}
\end{table*}

\begin{table*}
\centering
\scalebox{0.9}{
\begin{tabular}{cccccc}\\ 
\toprule  
 &Precision & Recall & F-1 measure & Fluency & Grammar \\ \midrule
\cite{Wang2018} &76.3&62.1&68.02& 2.98&3.06\\ 
Seq2Seq \cite{Bahdanau2015} &70.3&60.8&66.16&2.86&2.88\\
Pointer-Generator \cite{see2017} &76.6 &61.5 &67.95&\textbf{3.03}&3.02\\  
Structure-Aware Seq2Seq \cite{structure2018} &75.2 &61.7 &67.69&2.92&2.83\\  
\hline
Ours &\textbf{79.8}&\textbf{65.3} &\textbf{71.56}&3.01 &\textbf{3.10}\\
\bottomrule
\end{tabular}
}
\caption{Human Evaluation of various aspects of generated text.}
\label{table:human evaluation}
\end{table*}

\begin{figure}
    \centering
    \includegraphics[width=0.95\linewidth]{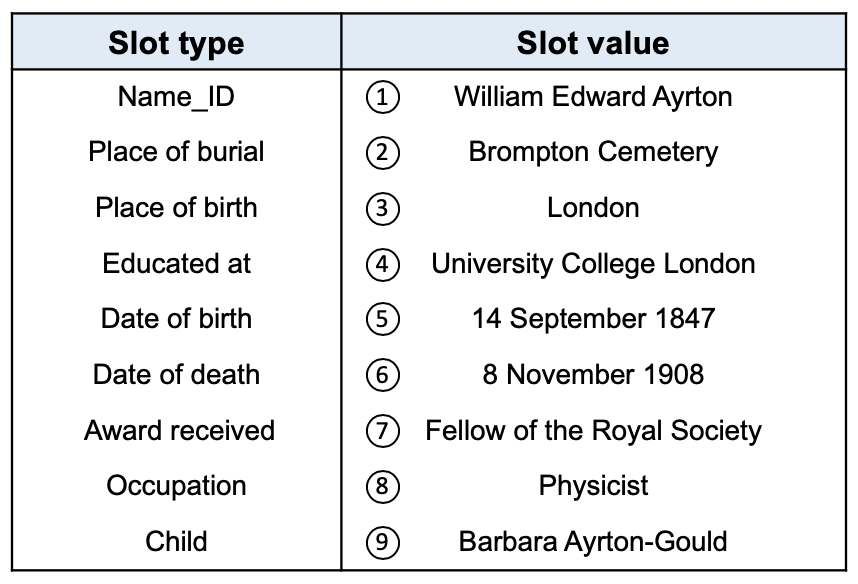}
    \caption{Example input for different models}
    \label{fig:case_table}
\end{figure}

\begin{table*}[t]
    \centering
    \begin{tabularx}{\textwidth}{l|p{2cm}|X}
         Model & Miss & Generated texts\\ 
         \hline
          \cite{Wang2018} &{  \textcolor{blue}{9}}& \small{William Edward Ayrton Fellow of the Royal Society ( 14 September 1847 -- 8 November 1908 ) was a British Physicist . Brompton Cemetery he was born in London\textcolor{red}{\textbf{ the son of Sir Thomas and his wife Mary ( n\'{e}e Fleming )}} . he was educated at \textcolor{red}{\textbf{University College School}} and University College London .}\\
         \hline
         Pointer generator &{ \textcolor{blue}{2},   \textcolor{blue}{9}}& \small{William Edward Ayrton-Gould Fellow of the Royal Society (14 September 1847 -- 8 November 1908) was an English Physicist who was born in London and was educated at \textcolor{red}{\textbf{Brompton College}} and University College London . he \textcolor{red}{\textbf{died in London}} on 8 November 1908 . William was elected a Fellow of the Royal Society  \textcolor{red}{\textbf{in 1902}}.} \\
         \hline
         Seq2Seq &{ \textcolor{blue}{1}, \textcolor{blue}{2},  \textcolor{blue}{3},   \textcolor{blue}{9}}& \small{William Edward \textcolor{red}{\textbf{Sandys}} Fellow of the Royal Society (14 September 1847 -- 8 November 1908) was a British Physicist . he was educated at the University College London and the University College London . he was a Fellow of the Royal Society and a Fellow of the Royal Society. }\\
         \hline
         Structure-Aware &{\textcolor{blue}{1}, \textcolor{blue}{2}, \textcolor{blue}{9}}& \small{William Edward \textcolor{red}{\textbf{Keeler}} Fellow of the Royal Society (14 September 1847 -- 8 November 1908) was a British Physicist and Physicist . he was elected a Fellow of the Royal Society \textcolor{red}{\textbf{in 1889}} and was a member of \textcolor{red}{\textbf{the Royal Society of London}} and \textcolor{red}{\textbf{the Royal Society of London}} and \textcolor{red}{\textbf{the Royal Society of London}} . he was educated at the University College London and at the University College London where he was \textcolor{red}{\textbf{a pupil of the chemist William}}.} \\
         \hline
         Ours & None& \small{William Edward Ayrton Fellow of the Royal Society (14 September 1847 -- 8 November 1908) was an English Physicist . William was born in London and educated at University College London. he is buried in Brompton Cemetery London . he was elected a Fellow of the Royal Society \textcolor{red}{\textbf{in 1901}}. he was the father of Barbara Ayrton-Gould .}\\
         \hline\hline
    \end{tabularx}
    \caption{Example outputs from different methods with an input table shown in Figure~\ref{fig:case_table}. The blue color indicates the corresponding row appears in the input table, but not in the output generation text. The red color indicates that these entities appear in the text but do not appear in the input table.}
    \label{tab:gen_example_1}
    \vspace{-0.4cm}
\end{table*}

\subsection{Baseline Models}

We compare our model with several strong baselines, including
\begin{itemize}
    \vspace{-0.2cm}
    \item The vanilla Seq2Seq attention model \cite{Bahdanau2015}.
    \vspace{-0.2cm}
    \item  The method in \cite{Wang2018}: The state-of-art model on the Wikiperson dataset. 
    \vspace{-0.2cm}
    \item The method in \cite{structure2018}: The state-of-the-art method on the WikiBio dataset.
    \vspace{-0.2cm}
    \item The pointer-generator \cite{see2017}: A Seq2Seq model with attention, copying and coverage mechanism.
\end{itemize}

\subsection{Implementation Details}
Our implementation is based on OpenNMT \cite{opennmt}. We train our models end-to-end to minimize our objective function with/without the copy mechanism.
The vocabulary is limited to the 50, 000 most common words in the training dataset. The hidden units of the multi-head component and the feed-forward layer are set to 2048. The baseline embedding size is 512. Following \cite{ALBERT2019}, the embedding size with embedding factorization is set to be 128. The number of heads is set to 8, and the number of Transformer blocks is 3. Beam size is set to be 5. Label smoothing is set to 0.1.

For the optimal-transport based regularizer, we first train the model without OT for about 20,000 steps, then fine tune the network with OT for about 10,000 steps. 
We use the Adam \cite{Kingma2015}
optimizer to train the models. We set the hyper-parameters of Adam optimizer accordingly, including the learning rate $\alpha$ = 0.00001, and the two momentum parameters, batch size = 4096 (tokens) and $\beta_2 = 0.998$.

\subsection{Results}
Table \ref{tab:baseline} and \ref{tab:PR} show the experiment results in terms of different evaluation metrics compared with different baselines. \quotes{Ours} means our proposed model with components of copy mechanism, embedding factorization, OT-matching with nouns, and latent similarity loss\footnote{The result of the method by \cite{Wang2018} is different from the score reported in their paper, as we use their publicly released code \href{url}{https://github.com/EagleW/Describing\_a\_Knowledge\_Base} and data that is three times larger than the original 106,216 table-text pair data used in the paper. We have confirmed the correctness of our results with the author.}. We can see that our model outperforms existing models in all of the automatic evaluation scores, indicating high quality of the generated texts. The superiority of the PARENT-T scores (in terms of precision and recall) indicates that the generated text from our model is more faithful than others. Example outputs from different models are shown in Table \ref{tab:gen_example_1} with an input table shown in Figure \ref{fig:case_table}. In this example, our model covers all the entities in the input, while all other models miss some entities. Furthermore, other models hallucinate some information that does not appear in the input, while our model generates almost no extra information other than that in the input. These results indicate the faithfulness of our model. More examples are shown in Appendix \ref{app:more_example}.

\subsection{Ablation Study}

We also conduct extensive ablation studies to better understand each component of our model, including the copy mechanism, embedding factorization, optimal transport constraint loss, and latent similarity loss. Table \ref{tab:ablation} shows the results in different evaluation metrics.

\vspace{-0.3cm}
\paragraph{Effect of copy mechanism}
The first and second rows in Table \ref{tab:ablation} demonstrate the impacts of the copy mechanism. It is observed that with the copy mechanism, one can significantly improve the performance in all of the automatic metrics, especially on the faithfulness reflected by the PARENT-T score.

\vspace{-0.1cm}
\paragraph{Effect of embedding factorization}
We compare our model with the one without embedding factorization. The comparisons are shown in the second and third rows of Table \ref{tab:ablation}. We can see that with embedding factorization, around half of the parameters can be reduced, while comparable performance can still be maintained.

\vspace{-0.1cm}
\paragraph{Effect of table-text embedding similarity loss}
We also test the model by removing the table-text embedding similarity loss component. The third and fourth rows in Table \ref{tab:ablation}  summarize the results. With the table-text embedding similarity loss, the BLEU and METEOR scores drop a little, but the PARENT and PARENT-T scores improve over the model without the loss. This is reasonable because the loss aims at improving faithfulness of generated texts, reflected by the PARENT-T score.

\vspace{-0.2cm}
\paragraph{Effect of the OT constraint loss}
We further compare the performance of the model (a) without using OT loss, (b) with using the whole table and text to compute OT, and (c) with using the extracted nouns from both table and text to compute OT. Results are presented in the third, fifth, and sixth rows of Table~\ref{tab:ablation}, respectively. The model with the OT loss improve performance on almost all scores, especially on the PARENT-T score. Furthermore, with only using the nouns to compute the OT loss, one can obtain even better results. These results demonstrate the effectiveness of the proposed OT loss on enforcing the model to be faithful to the original table.

\subsection{Human Evaluation}
Following \cite{Wang2018, tian2019}, we conduct extensive human evaluation on the generated descriptions and compare the results to the state-of-the-art methods. We design our evaluation criteria based on \cite{Wang2018, tian2019}, but our criteria differs from \cite{tian2019} in several aspects. Specifically, for each group of generated texts, we ask the human raters to evaluate the grammar, fluency, and faithfulness. The human evaluation metrics of faithfulness is defined in terms of precision, recall and F1-score with respect to the reconstructed Knowledge-base table from a generated text sequence. To ensure accurate human evaluation, the raters are trained with word instructions and text examples of the grading standard beforehand. During evaluation, we randomly sample 100 examples from the predictions of each model on the Wikiperson test set, and provide these examples to the raters for blind testing. More details about the human evaluation are provided in the Appendix \ref{app:human_evaluation}. The human evaluation results in Table \ref{table:human evaluation} clearly show the superiority of our proposed method.

\section{Related Work}
Table-to-text generation has been widely studied, and Seq2Seq models have achieved promising performance. \cite{Wikiemnlp2016,structure2018,Wiseman2018,Ma2019,Wang2018,tianyu2019}. 

For Transformer-based methods, the Seq2Seq Transformer is used by \citet{Ma2019} for table-to-text generation in low-resource scenario. Thus, instead of encoding an entire table as in our approach, only the predicted key facts are encoded in \cite{Ma2019}. Extended transformer has been applied to game summary \cite{gong2019} and E2E NLG tasks \cite{Gehrmann2018}. However, their goals focus on matching the reference text instead of being faithful to the input. 

Another line of work attempts to use external knowledge to improve the quality of generated text~\cite{chen2019}. These methods allow generation from an expanded external knowledge base that may contain information not relevant to the input table. Comparatively, our setting requires the generated text to be faithful to the input table. \citet{Nie2018} further study fidelity-data-to-text generation, where several executable symbolic operations are applied
to guide text generation. Both models do not consider the matching between the input and generated output.

Regarding datasets, most previous methods are trained and evaluated on much simpler datasets like WikiBio~\cite{Wikiemnlp2016} that contains only one sentence as a reference description. Instead, we focus on the more complicated structured knowledge base dataset \cite{Wang2018} that aims to generate multi-sentence texts. 
\citet{Wang2018} propose a model based on the pointer network that can copy facts directly from the input knowledge base. Our model uses a similar strategy but obtains much better performance.

In terms of faithfulness, one related parallel work is \citet{tian2019}. However, our method is completely different from theirs. Specifically, \citet{tian2019} develop a confidence oriented decoder that assigns a confidence score
to each target position to reduce the unfaithful information in the generated text. Comparatively, our method enforces faithfulness by including the proposed table-text optimal-transport matching loss and table-text embedding similarity loss.
Moreover, the faithfulness of \citet{tian2019} only requires generated texts to be supported by either a table or the reference; whereas ours constrains generated texts to be faithful only to the table.

Other related works are \cite{bootstrapgen2018, tabletotext2019}. For \cite{bootstrapgen2018}, the content selection mechanism training with multi-task learning and reinforcement learning is proposed. For \cite{tabletotext2019}, they propose force attention and reinforcement learning based method. Their learning methods are completely different from our method that simultaneously incorporates optimal-transport matching loss and embedding similarity loss. Moreover, the REINFORCE algorithm \cite{william1992} and policy gradient method used in \cite{bootstrapgen2018, tabletotext2019} exhibits high variance when training the model.

Finally, the content-matching constraints between text and table is inspired by ideas in machine translation \cite{sentagree2019} and Seq2Seq models \cite{liqun2019}.

\section{Conclusion}
In this paper, we propose a novel Transformer-based table-to-text generation framework to address the faithful text-generation problem. To enforce faithful generation, we propose a new table-text optimal-transport matching loss and a table-text embedding similarity loss. To evaluate the faithfulness of the generated texts, we further propose a new automatic evaluation metric specialized to the table-to-text generation problem. Extensive experiments are conducted to verify the proposed method. Both automatic and human evaluations show that our framework can significantly outperform the state-of-the-art methods.

\section*{Acknowledgements}

We sincerely thank all the reviewers for providing valuable feedback. We thank Linfeng Song, Dian Yu,  Wei-yun Ma, and Ruiyi Zhang for the helpful discussions. 

\nocite{song2018graph}
\bibliography{table2text}
\bibliographystyle{acl_natbib}

\newpage\clearpage\newpage
\appendix
\section{PARENT-T Metric} \label{app:parent-t}
PARENT-T evaluates each instance $(T^i, G^i)$ separately, by computing the precision and recall of generated text $G^i$ against table $T^i$. In other words, PARENT-T is a table-focused version of PARENT \cite{Dhingra2019}.

When computing precision, we want to check what fraction of the n-grams in $G_n^i$ are correct. We consider an n-gram $g$ to be correct if it has a high probability of being entailed by the table.
We use the word overlap model for entailment probability $w(g)$.
The precision score $E_p$ for one instance is computed as follows:
\begin{align}
    w(g) &= \frac{\sum_{j=1}^n \mathbb{1}(g_j\in \bar{T}^i)}{n} \\
    E_p^n &= \frac{\sum_{g\in G_n^i}w(g) \#_{G_n^i}(g)}{\sum_{g\in G_n^i}\#_{G_n^i}(g)} \\
    E_p &= \text{exp}\left(\sum_{n=1}^4 \frac{1}{4}\text{log}E_p^n\right)
\end{align}
where $\bar{T}^i$ denotes all the lexical items present in the table $T^i$, $n$ is the length of $g$, and $g_j$ is the $j$th token in $g$. $w(g)$ is the entailment probability, and $E_p^n$ is the entailed precision score for n-grams of order $n$. $\#_{G_n^i}(g)$ denotes the count of n-gram $g$ in $G_n^i$. The precision score $E_p$ is a combination of n-gram orders 1-4 using a geometric average.

For recall, we only compute it against table to ensure that texts that mention more information from the table get higher scores. 
$E_r(T^i)$ is computed in the same way as in \citet{Dhingra2019}:
\begin{equation}
    E_r = E_r(T^i) = \frac{1}{K}\sum_{k=1}^K\frac{1}{|\bar{r}_k|}LCS(\bar{r}_k, G^i)
\end{equation}
where a table is a set of records $T^i = \{r_k\}_{k=1}^K$, $\bar{r}_k$ denotes the value string of record $r_k$, and $LCS(x,y)$ is the length of the longest common subsequence between $x$ and $y$. Higher values of $E_r(T^i)$ denote that more records are likely to be mentioned in $G^i$.

Thus, the PARENT-T score (\textit{i.e.} F score) for one instance is:
\begin{equation}
    \text{PARENT-T} = \frac{2E_p E_r}{E_p + E_r}
\end{equation}
The system-level PARENT-T score for a model $M$ is the average of instance-level PARENT-T scores across the evaluation set.

\section{Details of Human Evaluation} \label{app:human_evaluation}

The following are the details for instructing our human evaluation raters how to rate each generated sentence:

We only provide the input table and the generated text for the raters. There are 20 well-trained raters participating in the evaluation.

\paragraph{Fluency}:

4: The sentence meaning is clear and flow naturally and smoothly.

3: The sentence meaning is clear, but there are a few interruptions.

2: The sentence does not flow smoothly but people can understand its meaning.

1: The sentence is not fluent at all and people cannot understand its meaning.

\paragraph{Grammar}:

4 : There are no grammar errors.

3:  There are a few grammar errors, but sentence meaning is clear.

2:  There are some grammar errors, but not influencing its meaning.

1:  There are many grammar errors. People cannot understand the sentence meaning.

\paragraph{Faithfulness}

A sentence is faithful if it contains only information supported by the table. It should not contain additional information other than the information provided by the table or inferred from the table. Also, the generated sentence should cover as much information in the given table as possible. The raters first manually extract entities from the generated sentences and then calculate the precision as the percentage of entities in the generated text also appear in the table; calculate the recall as the percentage of entities in the table also appear in the generated text. For each table-text pair, its F-1 score is then calculated according to the precision and recall.

\section{IPOT algorithm}\label{app:IPOT}

Given a pair of table and its corresponding text description, we can obtain table words embedding as $\bm{S} = \{\bm{x}_i\}_{i=1}^{i=n}$, and the model output for 
sentence words embedding as $\bm{S}^{\prime} = \{\bm{y}_j\}_{j=1}^{j=m}$. The cost matrix $\bm{C}$ is then computed as in Section \ref{sec:OT}. Both $\bm{S}$ and $\bm{S}^{\prime}$ are used as inputs to the IPOT algorithm in Algorithm~\ref{transitive} to obtain the OT-matching distance.

\begin{algorithm}[ht!]
	\caption{IPOT algorithm.}\label{transitive}
	\begin{algorithmic}
	\REQUIRE Feature vector $\bm{S} = \{\bm{x}_i\}_{i=1}^{i=n}$, $\bm{S}^{\prime} = \{\bm{y}_j\}_{j=1}^{j=m}$, and stepsize $1/\beta$
	
	\STATE $\bm{\sigma} = \frac{1}{m} \bm{1_m}$ $\bm{T}^{1} = \bm{1_n} \bm{1_m}^T$
	
	\STATE $\bm{C}_{ij} = d(\bm{x}_i, \bm{y}_j)$, $\bm{A}_{ij} = e^{-\frac{C_{ij}}{\beta}}$
	\FOR{$t = 1$ to $N$}
	   \STATE $\bm{Q} = \bm{A} \odot \bm{T}^{t}$
	   \FOR{$k = 1$ to $K$}
	      \STATE $\bm{\delta} = \frac{1}{n Q \bm{\sigma}}, \bm{\sigma} =\frac{1}{m Q^{T} \bm{\delta} }$
	   \ENDFOR
	   \STATE $\bm{T}^{t+1} = diag(\bm{\delta})\bm{Q} diag(\bm{\sigma})$
	\ENDFOR

	\RETURN $\bm{T}$

	\end{algorithmic}
\end{algorithm}

\section{Details of Optimal Transport Loss} \label{app:OTloss}
\begin{figure*}
    \centering
    
    \begin{minipage}{0.6\linewidth}
        \includegraphics[width=\linewidth]{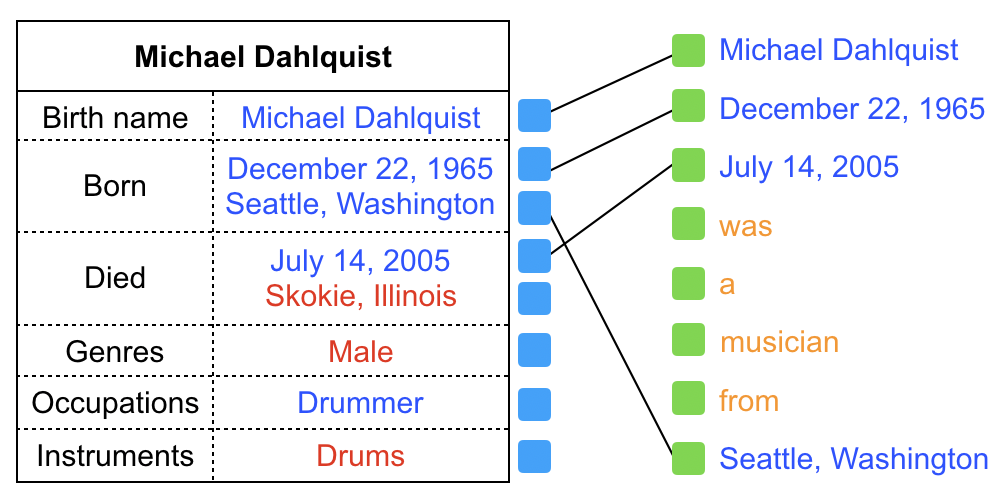}
    \end{minipage}
    \vfill
    \begin{minipage}{0.6\linewidth}
        \includegraphics[width=\linewidth]{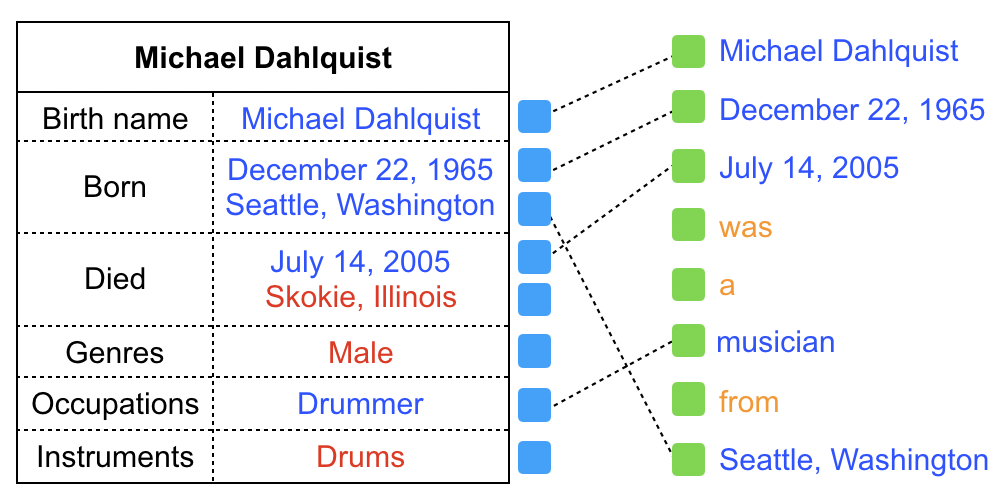}
    \end{minipage}
    \vfill 
    \begin{minipage}{0.6\linewidth}
        \includegraphics[width=\linewidth]{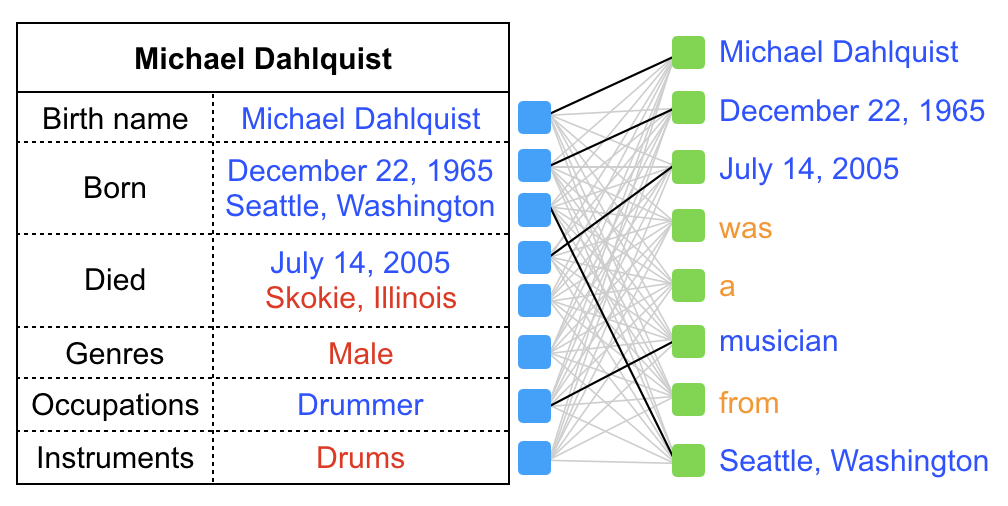}
    \end{minipage}
    
    \caption{
    Hard matching (top), soft bipartite matching (middle), and optimal transport matching (bottom).}\label{fig:OT_cases}
    \vspace{-0.4cm}
\end{figure*}

Figure \ref{fig:OT_cases} illustrates three matching cases from top to bottom, namely hard matching, soft bipartite matching, and optimal transport matching. The hard matching stands for exactly matching words between the table and the target sequences. This operation is non-differentiable. The soft bipartite matching, on the other hand, supposes the similarity between the word embedding $\mathbf{v}_{i_{k}}$ and $\mathbf{v}_{j_{k}}^{\prime}$ is $d(\mathbf{v}_{i_{k}}, \mathbf{v}_{j_{k}}^{\prime})$, and finds the matching such that $\mathcal{L} = \sum_{k} d(\mathbf{v}_{i_{k}}, \mathbf{v}_{j_{k}}^{\prime}) $ is minimized. This minimization can be solved exactly by the Hungarian algorithm \cite{kuhn1955}. But, its objective is still non-differentiable. Our proposed optimal transport matching can be viewed as the relaxed problem of the soft bipartite matching by computing the distance between the distribution over the input table and the decoded text sentence. This distance in optimal transport matching is differentiable.

\section{More generation examples} \label{app:more_example}
\begin{figure*}
    \centering
    \includegraphics[width=0.425\linewidth]{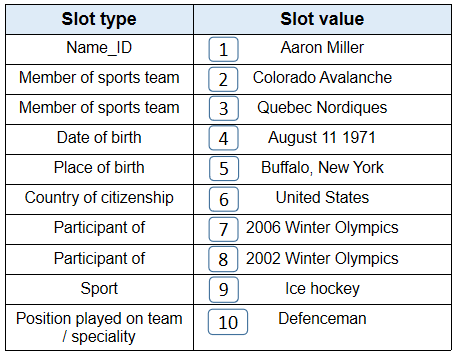}
    \caption{Example input for different models.}
    \label{figapp:case_table2}
\end{figure*}

\begin{figure*}
    \centering
    \includegraphics[width=0.95\linewidth]{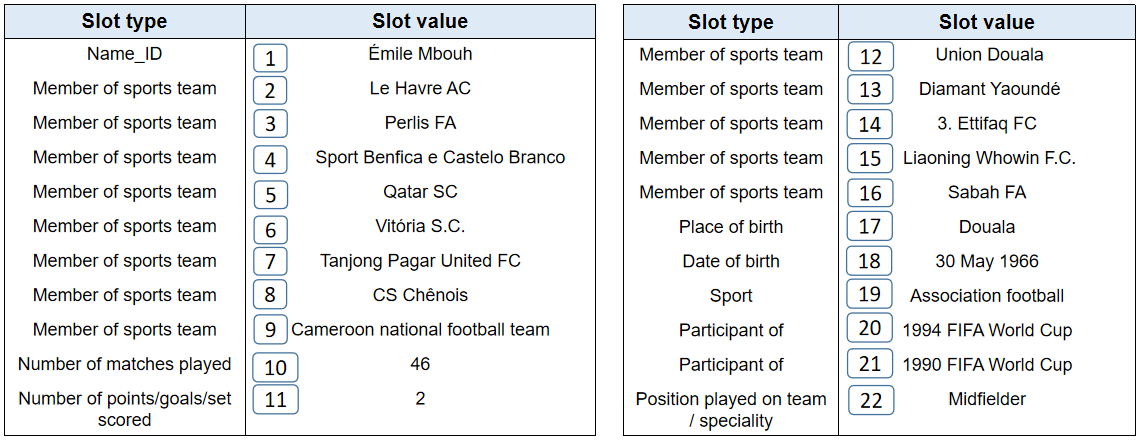}
    \caption{Example input for different models.}
    \label{figapp:case_table1}
\end{figure*}

More generation examples from different models are shown in Figure \ref{figapp:case_table2}, \ref{figapp:case_table1}, and Table  \ref{tabapp:gen_example_2}, \ref{tabapp:gen_example_1}.
Specifically, Table \ref{tabapp:gen_example_1} and Figure \ref{figapp:case_table1} show a more challenging example, as its table has 22 rows. In this example, we can observe that all the RNN-based models cannot capture such long term dependencies and miss most of the input records in the table. By contrast, our model miss much less input records.

\begin{table*}[t]
    \centering
    \begin{tabularx}{\textwidth}{l|p{2.3cm}|X}
    \toprule
         Model & Miss & Generated texts\\ 
         \hline
          \cite{Wang2018} & \small{7, 8}& \small{Aaron Miller ( born August 11 1971 ) is an United States former professional Ice hockey Defenceman who played in the National Hockey League ( NHL ) for the Quebec Nordiques and the Colorado Avalanche . he was born in Buffalo, New York and played for the Quebec Nordiques and the \textcolor{red}{\textbf{Ottawa Senators}} .}\\
         \hline
         Pointer generator & \small{2, 7, 8}& \small{Aaron Miller (born August 11 1971) is a retired United States professional Ice hockey Defenceman who played in the National Hockey League (NHL) for the Quebec Nordiques Quebec Nordiques Quebec Nordiques and the Quebec Nordiques . he was born in Buffalo, New York and \textcolor{red}{\textbf{grew up in New York City,}}} \\
         \hline
         Seq2Seq & \small{3, 7, 8}& \small{Aaron Miller (born August 11 1971) is an United States former professional Ice hockey Defenceman who played in the National Hockey League . miller was born in Buffalo, New York . \textcolor{red}{\textbf{he was drafted by the Colorado Avalanche in the second round of the 1994 NHL Entry Draft . he was drafted in the sixth round of the 1994 NHL Entry Draft by the Colorado Avalanche . he was drafted in the sixth round of the 1994 NHL Entry Draft by the Colorado Avalanche . }}}\\
         \hline
         Structure-Aware & \small{ 7, 8}& \small{Aaron Miller (born August 11 1971 in Buffalo, New York New York) is a retired United States professional Ice hockey Defenceman who played in the National Hockey League (NHL) for the Quebec Nordiques Colorado Avalanche Colorado Avalanche Colorado Avalanche Colorado Avalanche and Quebec Nordiques. \textcolor{red}{\textbf{he was drafted in the 2nd round of overall of the 2002 NHL Entry Draft.}}} \\
         \hline
         Ours & \small{None}& \small{Aaron Miller (born August 11 1971 in Buffalo, New York) is an United States former professional Ice hockey Defenceman who played in the National Hockey League (NHL) for the Quebec Nordiques and Colorado Avalanche . he was a member of the United States men's national Ice hockey team at the 2002 Winter Olympics and 2006 Winter Olympics.}\\
         \hline\hline
    \end{tabularx}
    \caption{Example outputs from different methods with an input table shown in Figure~\ref{figapp:case_table2}. The \quotes{Miss} column indicates the corresponding row appears in the input table, but does not appear in the output generation text. The red color indicates that these entities appear in the text but do not appear in the input table.}
    \label{tabapp:gen_example_2}
    \vspace{-0.3cm}
\end{table*}

\begin{table*}[t]
    \centering
    \begin{tabularx}{\textwidth}{l|p{2.3cm}|X}
    \toprule
         Model & Miss & Generated texts\\ 
         \hline
          \cite{Wang2018} & \small{2, 3, 4, 5, 6,  8, 10, 11, 12, 13, 14, 15, 16, 21, 22}& \small{\'{E}mile Mbouh ( born 30 May 1966 ) is a former Cameroon national football team Association football . he was born in Douala and played for the Tanjong Pagar United FC in the 1994 FIFA World Cup .}\\
         \hline
         Pointer generator & \small{2, 4, 5, 6, 7, 8, 10, 11, 12, 13, 14, 16, 16, 20}& \small{\'{E}mile Mbouh, (born 30 May 1966) is a Cameroon retired Association football who played as a Midfielder . he played for Cameroon national football team in the 1990 FIFA World Cup . he also played for Perlis FA and Liaoning Whowin F.C. . \'{E}mile was born in Douala,} \\
         \hline
         Seq2Seq & \small{2, 3, 4, 5, 6, 7, 8, 9, 10, 11, 12, 13, 14, 15, 16, 20}& \small{\'{E}mile Mbouh, (born 30 May 1966) is a retired Cameroonian Association football who played as a Midfielder . he was born in Douala . he was a member of the Cameroon national football team at the 1990 FIFA World Cup . he was a member of the Cameroon national football team at the 1990 FIFA World Cup . he was a member of the Cameroon national football team at the 1990 FIFA World Cup . he was a member of the Cameroon national football team at the 1990 FIFA World Cup .}\\
         \hline
         Structure-Aware & \small{2, 3, 4, 5, 6, 8, 10, 11, 12, 13, 14, 15, 16, 17, 21 }& \small{\'{E}mile Mbouh, (born 30 May 1966) is a Cameroonian retired Association football who played as a Midfielder . \textcolor{red}{\textbf{Le}} represented Cameroon national football team at the 1994 FIFA World Cup and 1994 FIFA World Cup . he played for \textcolor{red}{\textbf{Le FC Sport Yaound\'{e}}}, \textcolor{red}{\textbf{United Yaound\'{e}}} and Tanjong Pagar United FC} \\
         \hline
         Ours & \small{2, 3, 5, 6, 8, 12, 13, 14 }& \small{\'{E}mile Mbouh (born 30 May 1966) is a Cameroonian retired Association football who played as a Midfielder . born in Douala \'{E}mile began his career with Sport Benfica e Castelo Branco and Tanjong Pagar United FC . he also represented Cameroon national football team at the 1994 FIFA World Cup and 1990 FIFA World Cup . he also played for Sabah FA and Liaoning Whowin F.C. in the Malaysia Super League . he also played for Tanjong Pagar United FC and Liaoning Whowin F.C. in the Chinese Super League.}\\
         \hline\hline
    \end{tabularx}
    \caption{Example outputs from different models with an input table shown in Figure~\ref{figapp:case_table1}. The \quotes{Miss} column indicates the corresponding row appears in the input table, but does not appear in the output generation text. The red color indicates that these entities appear in the text but do not appear in the input table.}
    \label{tabapp:gen_example_1}
\end{table*}

\end{document}